\documentclass{article}
\usepackage{spconf,amsmath,graphicx}
\usepackage{colortbl}
\usepackage{amsmath}
\usepackage{hyperref}
\definecolor{Gray}{gray}{0.9}
\usepackage{comment} 
\usepackage{wrapfig}
\usepackage{subcaption} 

\title{Enhancing image quality and anomaly detection for small and dense industrial objects in nuclear recycling}

\begin{document} 
\maketitle
\begin{abstract}
This paper tackles two key challenges: detecting small, dense, and overlapping objects (a major hurdle in computer vision) and improving the quality of noisy images, especially those encountered in industrial environments. \cite{zhao2019object, zou2023object}. Our focus is on evaluating methods built on supervised deep learning. We perform an analysis of these methods, using a newly developed dataset comprising over 10k images and 120k instances. By evaluating their performance, accuracy, and computational efficiency, we identify the most reliable detection systems and highlight the specific challenges they address in industrial applications. This paper also examines the use of deep learning models to improve image quality in noisy industrial environments. We introduce a lightweight model based on a fully connected convolutional network. Additionally, we suggest potential future directions for further enhancing the effectiveness of the model.
\end{abstract}
\textit{The repository of the dataset and proposed model can be found at:  \href{https://github.com/o-messai/SDOOD}{https://github.com/o-messai/SDOOD}}, \\
\textit{\href{https://github.com/o-messai/DDSRNet}{https://github.com/o-messai/DDSRNet}}


\section{INTRODUCTION}
\label{sec:intro} 
\begin{figure}
    \centering
    \includegraphics[scale=0.120]{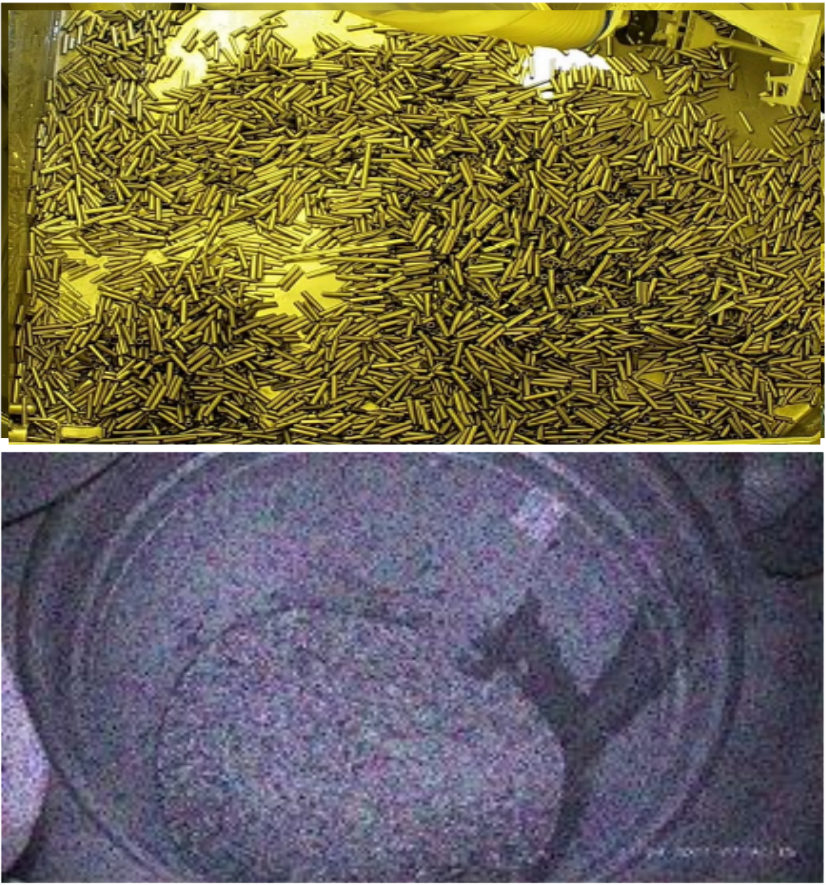}
    \includegraphics[scale=0.137]{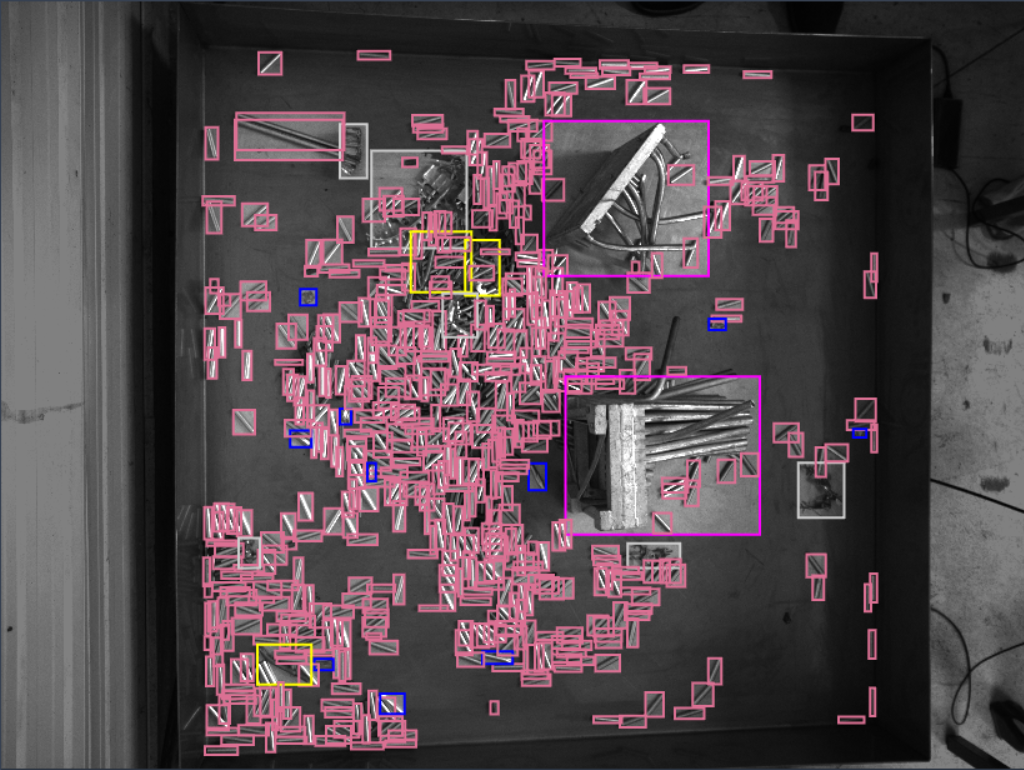}
    \caption{Top left: sorting table; bottom left: noisy image from the recycling process; right: simulation table image featuring annotated objects.}
    \label{fig:1}
\end{figure}

Object detection algorithms are a critical technology in many industrial applications, where they enable automation, quality control, and improved decision-making processes \cite{zhao2019object}. Although significant advances have been made in object detection algorithms through the application of deep learning techniques, the detection of small and overlapping objects remains a particularly challenging task \cite{rabbi2020small}. Industries such as manufacturing, robotics, logistics, and quality assurance, where precise identification and tracking of items are crucial for efficiency and safety, stand to benefit greatly from improved solutions to this problem. In these settings, the ability to accurately detect small objects that can be partially obfuscated or overlapped is crucial for tasks such as assembly line monitoring, inventory management, and defect detection. Detecting small, dense, and overlapping objects poses significant challenges in computer vision, as shown in Fig. \ref{fig:1} due to the complex nature of these scenes, where objects exhibit high variability in scale, position, and occlusion \cite{zou2023object}. Industrial applications often involve scenarios with objects that are not only small but also densely clustered and overlapping, making accurate detection both critical and difficult. In such environments, detection systems are prone to higher rates of false positives/negatives and missed detections. When objects overlap, parts of one or more objects may be obscured, making it difficult for the detection algorithm to accurately identify and delineate the individual objects. These challenges lie in the limitations of both traditional and machine-learning-based object detection algorithms, which often struggle with distinguishing features and maintaining accuracy under such conditions.

Real-time object detection has made remarkable strides with the advent of deep learning, ushering in a significant transition from traditional computer vision techniques to data-driven approaches. Recent studies in this field indicate ranging from the first hand-crafted feature-based techniques to modern convolutional neural network (CNN) \cite{ren2015faster} architectures such as YOLO \cite{Wang_2023_CVPR}, SSD, and Faster R-CNN \cite{ren2016faster}. Most of the proposed methods \cite{zou2023object} emphasize the progress made in detection accuracy and computational efficiency, while also acknowledging the persistent challenges in detecting small and overlapping objects, particularly in industrial environments. Emerging trends, such as multi-scale feature fusion and attention mechanisms, offer promising avenues for addressing these challenges and enhancing detection performance. Image denoising and quality enhancement on the other hand are pivotal areas of image processing, playing a crucial role in enhancing the reliability of visual data across various fields. Recent proposed image denoising/enhancement methods \cite{elad2023image} also show the shift from traditional filtering and optimization-based methods to advanced deep learning models, such as generative adversarial networks (GANs) and  diffusion models. However, they also highlight the challenge of achieving a balance between effective noise reduction and the preservation of fine structural details. Similar to trends in object detection models, the incorporation of attention mechanisms and multi-layer feature learning presents promising directions for advancing denoising methods. 

Orano Group is internationally recognized as a leading facility for nuclear fuel recycling. It is responsible for processing irradiated nuclear fuel from various power plants, extracting valuable materials such as uranium and plutonium, while ensuring the secure management and storage of radioactive waste. The recycling process at Orano is a vital aspect of the facility’s operations, and optimizing this process is crucial for the future of the nuclear energy industry. A significant challenge in the recycling process is the detection and measurement of metal shells intended for reuse. Additionally, the presence of radiation in the facility degrades the quality of camera feeds, making accurate object detection difficult. To overcome these challenges, we first propose a new model to tackle image quality and provide an analysis of state-of-the-art (SOTA) object detection methods on this unique industrial environment (see Fig. \ref{fig:1}). In this paper, we address these issues within the context of Orano's operational settings.

\section{Image Quality Enhancement}
\label{sec:Proposed} 
Uranium radiation primarily induces localized energy deposition in camera sensors, leading to increased pixel values or, in severe cases, permanent pixel damage. This damage can result in dead pixels that consistently output a fixed value. Simulating the effects of uranium radiation on camera sensors requires advanced noise models capable of capturing both the localized energy deposition and potential pixel degradation, including the formation of dead pixels. In this study, we selected two types of noise: shot noise (Poisson noise) \cite{hasinoff2021photon} and salt-and-pepper noise \cite{chan2005salt} due to their ability to approximate the characteristics of radiation-induced noise. These include fluctuations in pixel intensity and the potential for localized energy deposition, which can simulate the behavior of dead pixels.

Shot noise simulates the random electron excitation caused by radiation particles striking the camera sensor. This results in unpredictable pixel intensity fluctuations that follow a Poisson distribution:

\begin{equation}
I_n(x,y) = \begin{cases}  
\text{Poisson}(I(x,y)), & \text{with probability } P \\
I(x,y), & \text{otherwise}
\end{cases}
\end{equation}

where $I_n(x,y)$ represents the noisy pixel value, and $I(x,y)$ denotes the original pixel intensity, and $P$ indicates the probabilities of Poisson noise occurring. This statistical behavior is analogous to photon shot noise encountered in low-light imaging conditions. Conversely, salt-and-pepper noise replicates the effect of faulty pixels caused by radiation-induced damage. It produces random black-and-white pixel artifacts, mathematically represented as:

\begin{equation}
I_n(x,y) = \begin{cases} 
I_{\text{max}}, & \text{with probability } p_s \\ 
I_{\text{min}}, & \text{with probability } p_p \\ 
I(x,y), & \text{otherwise}
\end{cases}
\end{equation}

Here, $I_{\text{max}}$ and $I_{\text{min}}$ represent the maximum and minimum pixel intensities, respectively, while $p_s$ and $p_p$ indicate the probabilities of salt and pepper noise occurring. This model effectively mimics the appearance of dead pixels and other forms of sensor degradation due to radiation exposure.


\subsection{Deep Joint Denoising and Super-Resolution}
\label{sec:Enhancement} 
\begin{figure*}[ht]
    \centering
    \includegraphics[scale=0.26]{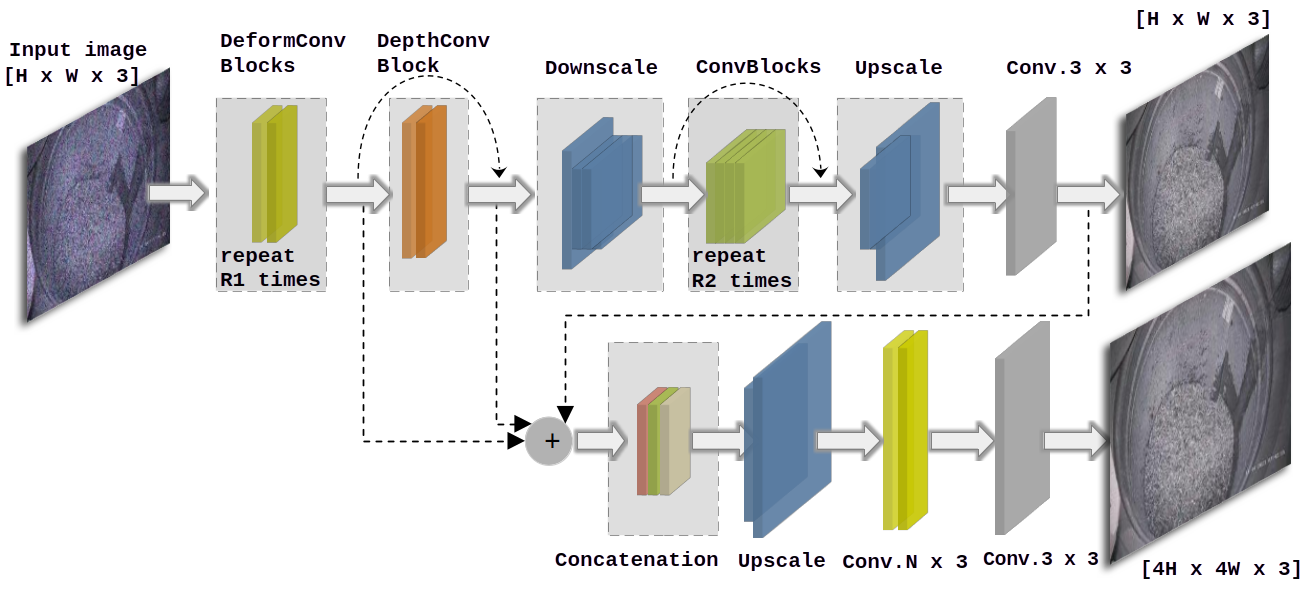}
    \caption{The proposed shallow model for image denoising and super-resolution.}
    \label{fig:enhancement_model}
\end{figure*}

We introduce a two-stage learning architecture DDSRNet (A Deep Model for Denoising and Super-Resolution) for simultaneous image denoising and super-resolution. By decoupling these challenging tasks, the model can specialize in each sub-task, leading to more effective noise removal and higher quality upscaling. The first stage focuses on denoising the input image, enabling the subsequent super-resolution stage to operate on cleaner data. This approach not only improves the robustness of the super-resolution process to noise but also allows the model to learn specialized noise removal strategies. To guide the model's learning, we propose a new loss function that jointly optimizes both denoising and super-resolution objectives, ensuring a balanced and effective solution. The model employs a series of convolutional layers with skip connections to process the input image, facilitating both denoising and super-resolution. As illustrated in Fig. \ref{fig:enhancement_model}, the network processes the input through multiple convolutional blocks, incorporating downscaling using $PixelUnshuffle$ and upscaling operations using $PixelShuffle$ to capture both high-frequency and low-frequency information. Specifically, the model utilizes a $DeformConv$ block followed by $DepthConv$ blocks and a $PixelUnshuffle$ block for downscaling. The extracted feature maps are then fed into $ConvBlock$ blocks, followed by $PixelShuffle$ for upscaling. Skip connections and concatenations are strategically implemented to combine feature maps from different layers, preserving information flow and enhancing the model's representational capacity. This design enables the model to generate both a denoised image and a super-resolution output.

The $DeformConv$ and $DepthConv$ blocks extract initial features, followed by subsequent layers that refine these features through additional convolutions and non-linear transformations. Downsampling is performed using $PixelUnshuffle$ to increase the channel dimension and reduce the spatial dimensions, while upsampling is achieved using $PixelShuffle$ to increase the spatial dimensions and reduce the channel dimension, crucial for the super-resolution task.  We consistently use a kernel size of $3x3$ for $DeformConv$ and $DepthConv$ blocks, while for the final convolution layer in the denoising and super-resolution paths, we set the kernel size to $1x1$. For the repeat block parameters, we set $R1=2$ and $R2=2$. More implementation details can be found in the source code. 

\subsection{Model training}
In order to minimize the error during training of the proposed model, we use the $L_{1}$ function combined with $PSNR$ and $SSIM$ loss functions as described as follows:
\begin{align}
\text{D} &= L_1(\text{I}_\text{ref} - \text{I}_\text{out}) 
           + 10 \cdot [1 - \text{SSIM}(\text{I}_\text{ref}, \text{I}_\text{out})] \notag \\
         &\quad + \frac{100}{\text{PSNR}(\text{I}_\text{ref}, \text{I}_\text{out})} \\
\text{Loss} &= \lambda \cdot \text{D}_\text{denoising} 
              + \beta \cdot \text{D}_\text{super-resolution}
\end{align}

where $I_{out}$ represents the denoised and super-resolved images, and $I_{ref}$ refers to the reference image. $L1$ computes the absolute difference between the reference and generated images. $SSIM$ (Structural similarity index measure) helps the model approximate the true structure of the reference, and $PSNR$ (Peak signal-to-noise ratio) guides the model toward improved fidelity. We employ a dynamic loss strategy where $\beta$, the super-resolution factor, is set to zero, and $\lambda$ is set to 1 for the first training epoch. This allows the model to focus on denoising initially. After each 10 epochs, we increase $\beta$ by 0.1 and decrease $\lambda$ by 0.1, stopping at epoch 50, where $\lambda$=$\beta$=0.5. This dynamic loss approach allows the model to initially focus on denoising and subsequently transition to super-resolution tasks. This method is better than static loss, as it allows for optimized training of both tasks. The model was trained and tested separately for each noise type.  For Salt and Pepper noise evaluation, each input image was corrupted by adding noise where the salt probability ($P_s$) and pepper probability ($P_p$) were randomly selected from range of [0-0.1]. For Shot noise, each image was similarly degraded, with the noise severity parameter $P$ from range of [0-0.2]. These ranges correspond to noise levels up to 10\% for each S\&P component and up to 20\% for Shot noise, respectively.

To update the weights of the model ($\approx$ 3 million parameters), we used the Stochastic Gradient Descent (SGD) with a momentum factor equals to 0.9, a weight decay factor sets to $10^{-4}$, a mini batch size equals to 128 and a learning rate initialized to $10^{-5}$. The Pytorch framework was used to implement our approach.

\subsection{Experimental Results}
\label{sec:Experimental_results}
We evaluate the performance of our proposed model on several well-known benchmark datasets for super-resolution, including Set5, Set14, and BSD100, along with our proprietary industrial dataset (UDD) (discussed in section \ref{sec:new_data}). Set5 \cite{zeyde2012single} consists of five uncompressed images featuring a baby, a bird, a butterfly, a head, and a woman. Set14 \cite{zeyde2012single} extends Set5 with nine additional images, offering a broader range of content that includes both natural and unnatural images. BSD100 \cite{arbelaez2010contour} contains 100 images, primarily depicting natural landscapes, food, and people. Finally, we select 100 images (80\% for train, and 20\% for test) from our 10k UDD dataset for evaluation. For noise simulations, we added two types of noise—Salt \& Pepper (S\&P) and Shot noise—to all datasets. The quantitative performance is measured using Peak Signal-to-Noise Ratio (PSNR) and Structural Similarity Index Measure (SSIM), which are standard metrics for evaluating image quality.

\begin{figure}[ht!]
    \centering
    \begin{minipage}{0.42\textwidth}
        \centering
        \resizebox{\linewidth}{!}{%
            \begin{tabular}{|c|c|c|c|c|}
                \hline
                Dataset & \multicolumn{2}{c|}{S\&P} & \multicolumn{2}{c|}{Shot noise} \\
                \cline{2-5}
                        & PSNR & SSIM & PSNR & SSIM \\
                \hline
                Set5 \cite{zeyde2012single}   & 28.17    & 0.83    & 27.24    & 0.82    \\
                Set14 \cite{zeyde2012single}  & 24.15    & 0.74    & 24.10    & 0.71    \\
                BSD100  \cite{arbelaez2010contour}  & 24.64    & 0.71    & 23.44    & 0.70    \\
                UDD     & 25.45    & 0.78    & 24.18    & 0.74    \\
                \hline
            \end{tabular}
        }
        \captionof{table}{Performance of the proposed DDSRNet on various datasets for 4× super-resolution under S\&P noise and Shot noise degradations.}
        \label{tab:measure_error}
    \end{minipage}
    \hfill
    \begin{minipage}{0.55\textwidth}
       
        \includegraphics[width=0.9\linewidth]{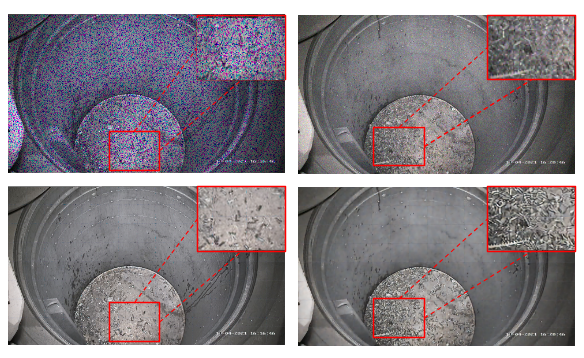}
        \caption{Visual image enhancement examples from \\our UDD dataset. Top: Noisy inputs. \\Bottom: Enhanced outputs. Left: S\&P noise. Right: Shot noise.}
        \label{fig:Enhancement_examples}
    \end{minipage}
\end{figure}

Table \ref{tab:measure_error} summarizes the results of our model across the four datasets under the presence of S\&P and Shot noise. For the Set5 dataset, our model achieves a PSNR of 28 dB and an SSIM of 0.83 under S\&P noise, while for Shot noise, it achieves a PSNR of 27 dB and an SSIM of 0.82. Similar trends are observed for the Set14 and BSD100 datasets, where the model consistently maintains competitive performance. Notably, our model demonstrates robust performance on the UDD dataset, achieving a PSNR of 25 dB and an SSIM of 0.78 under S\&P noise and 24 dB and 0.74 under Shot noise. The proposed model shows better performance under S\&P noise due to its ability to accurately identify and correct isolated pixel-level disturbances. This is enabled by its convolutional spatial filtering capabilities, which are particularly effective in detecting and restoring small, sparse noise patterns characteristic of S\&P noise, in contrast to the more diffuse and random distribution of Shot noise.
As illustrated in Fig. \ref{fig:Enhancement_examples}, the results demonstrate the ability of the model to effectively reconstruct high-quality images despite the presence of significant noise. Across all datasets, the PSNR and SSIM values indicate that the model can successfully mitigate noise while preserving fine details, demonstrating its generalization capability across different types of data. Beyond improving the visual quality for human operators in industrial settings, the proposed model enhances the performance of downstream object detection tasks. By denoising and super-resolving the input images, it provides clearer visual information, facilitating more accurate object recognition and classification. This dual benefit of image enhancement and noise reduction makes our approach highly valuable for industrial applications where both human and automated visual analysis are critical.

\section{Object Detection}
\label{sec:Object_detection} 
\subsection{New dataset}
\label{sec:new_data} 
The dataset for this study was collected using a simulation model developed in accordance with recommendations from Orano group. The simulated environments replicate real-world conditions as closely as possible. To account for variations in lighting, two types of light projection were used during image capture. The dataset consists of 10,000 images, which were meticulously labeled both manually and semi-automatically. In total, over 120,000 object annotations (bounding boxes) were made, with each image containing an average of 12 objects. The diversity of object shapes, sizes, and overlap within these images presents a particularly challenging environment for object detection models \cite{rabbi2020small}. Synthetic datasets offer scalable control over object properties and automated labeling. We generated a 50,000-image, 6.5-million-instance dataset using the open source Blender 3D software. However, evaluation on this synthetic data is beyond the current scope.

\subsection{Experimental Results}
\label{sec:Results} 
\begin{table*}[ht]
\centering
\resizebox{1.6\columnwidth}{!}{%
\begin{tabular}{|c|c|c||c|c|c|c|c|c|}
\hline
    Model & Img size  &$mAP0.5-0.95_{COCO-17}$ & mAP@0.5 & mAP0.5-0.95 & Recall& Precision & Parameters \\
\hline                     
         YOLOv5x \cite{jocher2022ultralytics} &$640^{2}$  &50.7 &42.6 &32.0 &54.6 &56.7  & 86.7M\\
         YOLOv6l \cite{li2022yolov6} &$640^{2}$   &52.8  &49.8 &39.2 &55.6 &53.7 & 59.6M\\  
         YOLOv7x \cite{Wang_2023_CVPR} &$640^{2}$  &53.1  &53.1 &38.0 &50.9 &87.4  & 71.3M\\ 
         YOLOv8m \cite{terven2023comprehensive} &$640^{2}$  &50.2  &40.6 &26.6 &34.0 & 57.9 & 25.9M\\           
         YOLOv8l \cite{terven2023comprehensive} &$640^{2}$  &52.9  &59.9 &45.0 &55.9 &80.6  & 43.7M\\ 
         YOLOv8x \cite{terven2023comprehensive} &$640^{2}$  &53.9  &60.8 &45.6 &56.8 &80.7 &68.2M \\                    
         YOLOv7-E6E \cite{Wang_2023_CVPR}  &$640^{2}$  &NA  & 51.3 & 36.3 & 49.6 &81.4 & 151.7M\\
         Yolo-nas-l \cite{zou2023object} &$640^{2}$  &52.2  & 50.9&NA &62.7  &14.3 & NA\\
         YOLOv9-e  \cite{wang2024yolov9} &$640^{2}$  &55.6 &57.9   &43.5 &51.9    &81.7   & 58.1M\\ 
         YOLOv10-x \cite{wang2024yolov10} &$640^{2}$  &54.4 & 48.3   &36.3 &48.5    &73.1   & 29.5M\\ 
         YOLO11-x \cite{yolo11_ultralytics} &$640^{2}$  &54.7  &\textbf{62.8}   &\textbf{46.7}  & 57.9   &81.9    &56.9M   \\ 
         Faster R-CNN-R101  \cite{ren2015faster} &$640^{2}$  &44.0 &50.1 &39.4 &61.2 &74.5 & 60M \\ 
         RT-DETR-l \cite{zhao2024detrs} &$640^{2}$  &53.0  &40.1 &27.9  &44.3 &69.1 & 32M \\
         RT-DETR-x \cite{zhao2024detrs} &$640^{2}$  &54.8  &48.9 &33.5 &50.1 &62.6  & 67M\\  
         \hline
         YOLOv8x \cite{terven2023comprehensive} &$1280^{2}$   &NA  &\textbf{65.4} &50.8 & 61.0 &68.8 &68.2M  \\ 
         YOLOv5x6 \cite{jocher2022ultralytics} &$1280^{2}$   &55.0  &44.1 &34.0   &52.0 &69.8 & 140.7M\\
         YOLOv6-L6 \cite{li2022yolov6} &$1280^{2}$  &57.2  &53.0 & 42.6 &50.9 &81.6& 140.4M \\    
         YOLOv9-e \cite{wang2024yolov9} &$1280^{2}$ &NA  &59.0   &47.0  &52.3   &81.3   &58.1M  \\ 
         YOLOv10-x \cite{wang2024yolov10}  &$1280^{2}$   &NA  &52.5   &42.6  & 49.2   &85.0   &29.5M  \\ 
         RT-DETR-l \cite{zhao2024detrs} &$1280^{2}$  &NA  &46.0 &33.5  &46.2 &69.2  & 32M \\ 
         RT-DETR-x \cite{zhao2024detrs} &$1280^{2}$  &NA  &50.9 & 35.7   &51.4  &63.6    & 67M \\ 
         YOLO11-x \cite{yolo11_ultralytics} &$1280^{2}$  &NA  &57.8   &44.1  & 60.3   &72.0   &56.9M  \\  

\hline
\end{tabular}
}
\caption{The models tested on our Orano UDD dataset (80\% train, and 20\% test) as well as their performances on COCO Microsoft. 
}
\label{tab:1}
\end{table*}

The performance of object detection models can be influenced by various factors, including convolution techniques, image resolution, and hyper-parameters employed during training. To determine the most effective model, we conducted a series of rigorous tests, summarized in Table \ref{tab:1}. The primary evaluation metric used was mean Average Precision (mAP) \cite{zou2023object}, which provides a comprehensive measure of model performance across multiple object scales and overlap conditions.

Table \ref{tab:1} presents a comparative analysis of various models across two image sizes, along with associated parameters. The models evaluated include several YOLO versions (YOLOv5, YOLOv6, YOLOv7, YOLOv8, YOLOv9, YOLO-NAS, YOLOv9 and YOLO11), Faster R-CNN-R101, and RT-DETR variants. The models are tested with images of sizes $640^{2}$ and $1280^{2}$, which allows for a broader understanding of performance scaling with input resolution. The results highlight that the YOLO family of models, particularly YOLOv8-x and YOLOv11-x, consistently exhibit state-of-the-art performance, demonstrating outstanding mAP scores and precision. These models effectively balance the trade-off between model size (number of parameters) and performance, optimizing them for real-world applications where both accuracy and efficiency are paramount. The superiority of YOLOv8 can be attributed to its use of an enhanced feature pyramid network (FPN) and path aggregation network (PAN), which effectively capture features across multiple scales, thereby improving the detection of small objects. This study emphasizes the need for innovative models specifically designed to meet the demands of such applications, including those in industrial environments, manufacturing, and recycling operations.

\subsection{Length measurement of objects}
\label{sec:length_measurement}
The usual methods for estimating object lengths in images involve employing a segmentation model such as \cite{kirillov2023segment} to extract object masks. These masks can then be used to measure pixel-based lengths, which can subsequently be converted to physical units such millimeters or centimeters. However, besides its good performance, this approach is computationally intensive and may exhibit sub-optimal performance when dealing with small, densely overlapping objects. As a result, we opt to utilize the diagonal of bounding boxes as a proxy for hull length. This approach offers a balance of speed, computational efficiency. First, the object is enclosed within a bounding box, defined by its top-left corner at \((x_{\text{min}}, y_{\text{min}})\) and bottom-right corner at \((x_{\text{max}}, y_{\text{max}})\). The diagonal of this bounding box can be computed using the Euclidean distance formula:
$d = \sqrt{(x_{\text{max}} - x_{\text{min}})^2 + (y_{\text{max}} - y_{\text{min}})^2}$
This diagonal length \(d\) gives a quick approximation of the length of the objects including metallic hull. The diagonal provides an estimate of the object's length for many shapes, as the length of most objects is roughly proportional to the diagonal of their bounding box. However, while this approach is fast, it is not always the most accurate. For irregularly shaped objects, the bounding box diagonal may not closely match the actual length of the object. In such cases, a more precise method involves object segmentation.


\begin{figure}[ht!]
    \centering
    \begin{minipage}{0.58\textwidth}
\resizebox{0.8\columnwidth}{!}{%
\begin{tabular}{|c|c|c|c|c|}
\hline
    Length in mm & Number of reps & TP &Detection accuracy \%  & Measurement Mean Error (+/- mm) \\
\hline                     

         65  & 30 &7 &23,33 &1   \\              
         70 & 30 &9 &30,00&3   \\              
         75 & 30 &15	&50,00 &9   \\              
         90 & 30 &13	 &43,33&4   \\              
         100 & 30 &18 &60,00 &11   \\              
         110 & 30 &23 &76,67&4   \\              
         120 & 30 &28 &93,33 &5   \\              
         130 & 30 &24 &80,00&7   \\              
         140 & 30 &26 &86,67 &7   \\              
         150 & 30 &21 &70,00 &10   \\              
         160 & 30 &24 &80,00 &10   \\              
         170 & 30 &27 &90,00 &11   \\              
         180 & 30 &25 &83,33 &11   \\              
         190 & 30 &27 &90,00 &16   \\              
         200 & 30 &22 &73,33 &20  \\              
         300 & 30 &27 &90,00 &98   \\              
         400 & 30 &27 &90,00 &210   \\              
         600 & 30 &28 &93,33 &377   \\              
        \hline
        \end{tabular}}
        \caption{Anomaly detection and measurement accuracy \\on a stack of 2k objects (metallic hulls of length 60 mm), \\conducting 30 random repetitions for each length category.}
        \label{tab:mesure_error}
    \end{minipage}%
    \hfill
    \begin{minipage}{0.4\textwidth}
        \centering
        \includegraphics[width=\linewidth]{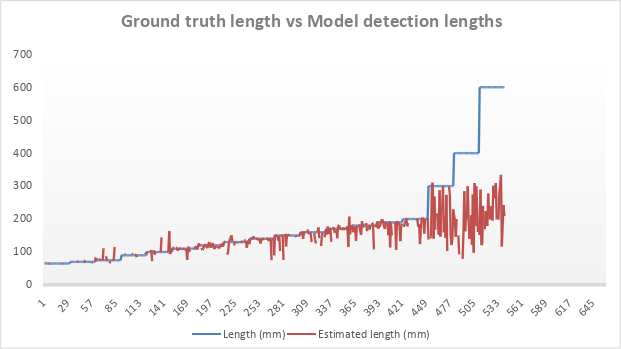}
        \caption{Identify and quantify long metallic hulls within a collection of 2k non-long metallic hulls of length 60 mm through (classified as "normal"). Conducting 30 random repetitions for each length category.}
        \label{fig:s4}
    \end{minipage}
\end{figure}

\begin{figure*}[ht]
    \begin{subfigure}[b]{0.5\linewidth}
        \centering
        \includegraphics[width=\linewidth]{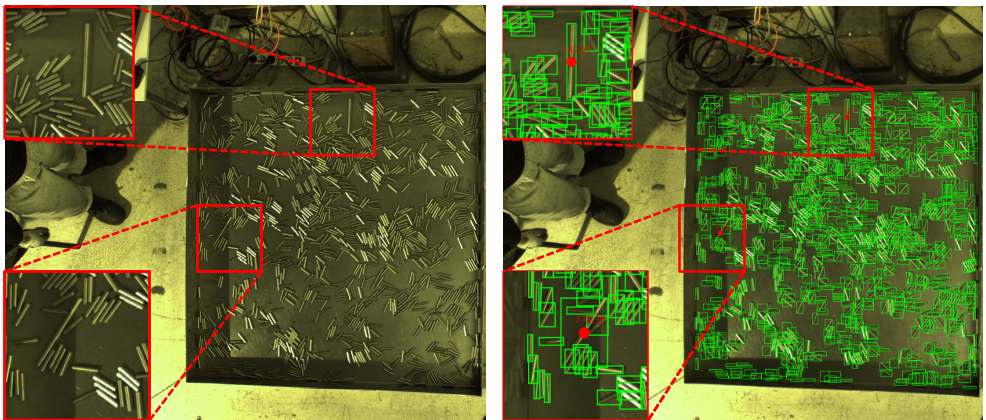}
        \caption{Anomaly detection with IQR over bounding box diagonals. The left image shows the input, while the right image shows the model's output.}
        \label{fig:anomaly_example}
    \end{subfigure}
    \hfill
    \centering
    \begin{subfigure}[b]{0.45\linewidth}
        \centering
        \includegraphics[width=\linewidth]{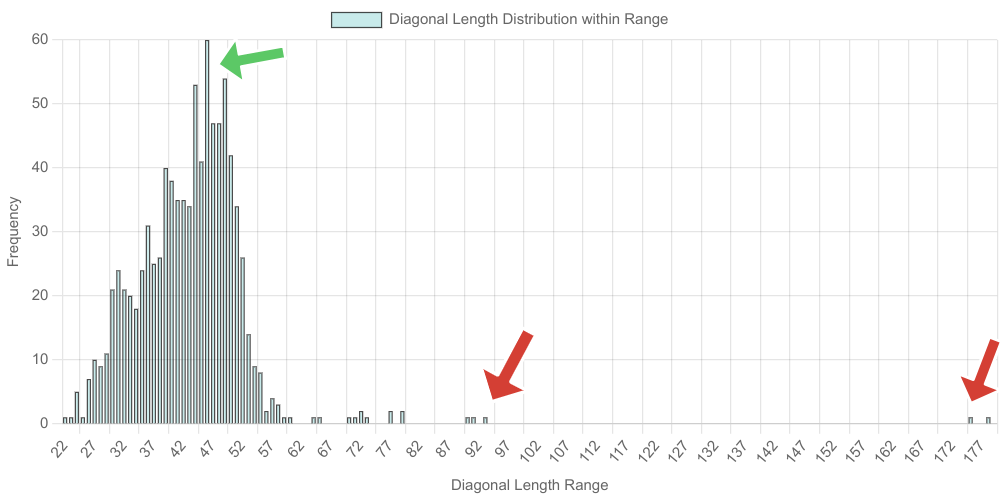}
        \caption{Outliers example in the bounding box diagonal distribution. Green arrows show the median, whereas red arrows indicate outliers.}
        \label{fig:outliers}
    \end{subfigure}
    \caption{Side-by-side figures showing outlier detection and anomaly detection examples.}
    \label{fig:side_by_side}
\end{figure*}

On-site evaluation has been performed, we assessed the model's capability to detect 1 to 4 long metallic hulls (e.i., "normal" or class 0) of varying lengths, ranging from 65 mm to 1100 mm. These hulls were introduced into a stack of 2000 hulls of length 60 mm through a randomized process, one at a time. The overall detection rate for long hulls (anomalies) reached 72\%. Table \ref{tab:mesure_error} provides a detailed breakdown of the detection performance for each hull length, including the associated measurement error in millimeters (following pixel calibration). Fig. \ref{fig:s4} highlights the model's limitations in accurately detecting long hulls or occasionally misclassifying them as multiple hulls. This behavior can be attributed to the imbalanced distribution of lengths within the training dataset. As we encounter hulls exceeding 300 mm, the model exhibits a tendency to detect them as two or three distinct hulls.

\subsection{Anomaly detection}
\label{sec:anomaly_detection}

To identify anomalies within the recycling line, specifically large objects and long metallic hulls, we employ straightforward statistical metric known as the Interquartile Range (IQR). 
This approach effectively highlights outliers by focusing on the central 50\% of the distribution, thereby demonstrating resilience to extreme values. A key advantage of using IQR is its independence from camera calibration or reference points, making it robust even when faced with changes in camera zoom or image resolution. Furthermore, this approach eliminates the necessity for explicit length measurements to identify long hulls or larger objects. The following outlines the procedure for identifying outliers using IQR:

First, the first quartile (Q1) is the 25th percentile, meaning 25\% of the data points are below this value, and the third quartile (Q3) is the 75th percentile, meaning 75\% of the data points are below this value. The Interquartile Range (IQR) is calculated as: $\text{IQR} = Q3 - Q1$. This gives the spread of the middle 50\% of the data. Next, the outlier boundaries are defined as follows:
$\text{Lower bound} = Q1 - 1.5 \times \text{IQR}$
$\text{Upper bound} = Q3 + 1.5 \times \text{IQR}$
Data points are flagged as outliers if they fall outside these bounds:
$\text{Data point} < Q1 - 1.5 \times \text{IQR}$ 
or
$\text{Data point} > Q3 + 1.5 \times \text{IQR}$
. This method is effective for outlier detection because it is robust to extreme values, as it relies on quartiles rather than the mean or standard deviation. However, given that our application focuses on larger and elongated objects, we exclusively employ upper-bound outliers as indicators of anomalies. Fig. \ref{fig:outliers} illustrates the distribution of bounding box diagonal lengths, with outliers identified using the IQR method. Furthermore, Fig. \ref{fig:anomaly_example} shows an example of finding long anomaly hulls without the need of length measurement/estimation.

This cost-effective method enables precise real-time anomaly detection, achieving up to 93\% accuracy even in harsh conditions, as demonstrated in Section \ref{sec:length_measurement}.

\section{Conclusion and perspectives}
\label{sec:conclusion_and_perspectives}
A new dataset has been created to address the challenge of detecting small, dense, and overlapping objects in industrial environments. The results from SOTA object detection methods are promising. The YOLOv8 and YOLOv11 series, particularly the -x variants, demonstrate the best trade-off between recall, precision, and parameter count, making them the top performers. However, there remains room for improvement and a need for more specialized model architectures. Furthermore, we proposed DDSRNet, a new denoising model that also enhances the resolution of input images. The shallow model is optimized for real-time processing to match the speed required for industrial recycling workflows. DDSRNet effectively mitigates the effects of radiation-induced noise, which significantly impacts image quality in industrial conditions. 
Overall, our work contributes to advancing industrial recycling processes and the broader field of object detection by offering robust solutions for challenging imaging environments and enhancing both human and automated analysis capabilities.



\bibliography{refs} 
\bibliographystyle{IEEEbib} 

\end{document}